# Exploring the Reversal Curse and Other Deductive Logical Reasoning in BERT and GPT-Based Large Language Models

Da Wu[1,2,*], Jingye Yang[1,2,*],  Kai Wang[1,3,**]

[1] Raymond G. Perelman Center for Cellular and Molecular Therapeutics, Children's Hospital of Philadelphia, Philadelphia, PA 19104, USA

[2] Department of Mathematics, University of Pennsylvania, Philadelphia, PA 19104, USA

[3] Department of Pathology and Laboratory Medicine, University of Pennsylvania, Philadelphia, PA 19104, USA

*: These authors contributed equally to this work.

**: Corresponding author and lead contact. Email: wangk@chop.edu

**SUMMARY**

The "Reversal Curse" describes the inability of autoregressive decoder large language models (LLMs) to deduce "B is A" from "A is B," assuming that B and A are distinct and can be uniquely identified from each other. This logical failure suggests limitations in using GPT models for tasks like constructing knowledge graphs. Our study revealed that the bidirectional LLM, BERT, does not suffer from this issue. To investigate further, we focused on more complex deductive reasoning by training encoder and decoder LLMs to perform union and intersection operations on sets. While both types of models managed tasks involving two sets, they struggled with operations involving three sets. Our findings underscore the differences between encoder and decoder models in handling logical reasoning. Thus, selecting BERT or GPT should depend on the task's specific needs, utilizing BERT's bidirectional context comprehension or GPT's sequence prediction strengths.

**INTRODUCTION**

Large Language Models (LLMs), such as BERT[1], GPT-4[2] and Llama 3[3], are advanced artificial intelligence (AI) models adept at understanding and producing human-like texts. Introduced in 2018, BERT (Bidirectional Encoder Representations from Transformers) revolutionized NLP with its ability to read text bidirectionally, greatly enhancing its understanding of language context. Meanwhile, the GPT (Generative Pretrained Transformer) series has advanced through iterations, each more adept than the last at generating coherent and contextually relevant texts, from answering questions to composing poetry. The most renowned closed-source foundational model at the time of writing this paper is GPT-4. Developed by OpenAI, GPT-4 is the latest iteration in the Generative Pre-trained Transformer series of language models. It marks a significant advancement over its predecessor, GPT-3[4], with enhanced capabilities in natural language understanding, generation, and context comprehension. GPT-4 leverages a larger and more diverse training dataset, enabling it to produce more accurate, coherent, and contextually relevant text. It is versatile, supporting applications such as text completion, translation, summarization, and conversational AI. Similarly, the most renowned open-source foundational model at the time of writing this paper is Llama 3. Developed with an emphasis on accessibility and transparency, Llama 3 offers similar functionality to GPT-4. It supports a wide range of applications, including text completion, translation, summarization, and conversational AI. Its open-source nature allows researchers and developers to fine-tune and integrate the model with various applications, making it a powerful and versatile tool in the field of artificial intelligence. Together, these LLMs demonstrate extraordinary capabilities in both comprehending and generating text, proving essential in advancing AI interaction.

The deductive logical reasoning capabilities of LLMs are crucial for many tasks and applications. These capabilities involve the process of reasoning from one or more statements (premises) to reach a more complex logically certain conclusion. This capability is fundamental in various intellectual activities, such as mathematical proofs, scientific reasoning, and logical augmentation. It is valued for its precision and certainty, as it provides conclusive results when the premises are accurate, and the reasoning process is correctly applied. The most basic task in deductive logic involves inferring "B is A" from "A is B". Here we assume that B and A are distinct and can be uniquely identified from each other, meaning that there are no alternative answers.

The term "Reversal Curse"[5] pertains to the observation that when autoregressive large language models (LLMs), such as GPT-3[6] and Llama-1[7], are trained on logical sentences structured as "A is B", they do not inherently grasp that "B is A" holds true as well. For instance, if the model was trained on the sentence "Jimmy Carter is the 39th president of the United States", then it will not correctly complete the sentence of "The 39th president of the United States is [ ]". **Figure 1** illustrates another example where GPT-4 (December 2023) knows that "Leonhard Euler's mother was Marguerite Brucker" but cannot answer the question "Who is Marguerite Brucker's son?" Therefore, many autoregressive LLMs fail to perform the simplest deductive reasoning, let alone the harder ones. Nevertheless, this issue can be addressed by in-context learning and logical deduction if the correct answer is already known beforehand.

The work by Grosse *et al.*[8] provide further evidence supporting the phenomenon known as the "Reversal Curse" through the use of influence functions. Their research indicates that a pretrained model might struggle with effective generalization when facts are not trained in both directions. Meng et al.[9] further support this, as they used a model editing approach to change factual associations and found that this technique does not work bidirectionally. Interestingly, the "Reversal Curse" also occurs in humans, where recalling information is more challenging in the reverse direction compared to the forward direction[10-14]. Several studies by Geva et al.[15-17] delve into the underlying processes of factual recall in Transformers, proposing that these models encode factual associations as key-value pairs within their feed-forward layers. This key-value storage method might partially explain the "reversal curse". From a slightly different angle, we aim to provide an alternative, more intuitive and less rigorous interpretation of the "reversal curse," drawing inspiration from the linear regression analysis above. In the training process of an auto-regressive decoder, a technique known as "masked self-attention" is employed[18]. This mechanism controls which tokens in the input can interact with other tokens, thereby influencing the model's training. More recent work by Mizrahi et al.[19] has introduced comprehensive benchmarks and prompt-specific evaluations, including assessments of logical reasoning and other measures. Other innovative methods for evaluating the reasoning abilities of large language models (LLMs) include structured debates[20]. The scope of research extends across various domains, highlighting logical deduction capabilities within fields such as marketing[21], mathematics and statistics[22-26], economics[27], law[28] and broader areas of scientific writing and formula generation[29,30].

However, the research mentioned above primarily focuses on autoregressive decoder language models. Consequently, our initial objective is to investigate these phenomena in bidirectional encoder language models, such as BERT (Bidirectional Encoder Representations from Transformers)[1], and to compare their performance with autoregressive decoder language models, such as GPT-4[2], Llama 2[31] and the recently introduced Llama 3[3]. In addition, we aim to delve deeper by assessing whether BERT and GPT models are capable of mastering intersection (∩) and union (∪) operations on two sets (through examples), and accurately executing these tasks in more intricate situations. Furthermore, we seek to comprehend the reasoning and dependability of BERT and GPT models in making correct predictions.

One driving force of this research is to support and offer valuable insights into the current endeavors of building medical knowledge graphs using LLMs[31-33], particularly focusing on autoregressive decoder models. Constructing an extensive knowledge graph from vast medical databases is an exceptionally challenging and intricate task, holding significant importance. In this context, the most crucial role of LLMs lies in their ability to infer relationships between various entities within existing database. This critical process involves the complex task of taking unions and intersections of multiple sets. Therefore, the current study evaluates the following tasks involving deductive reasoning:

1) Using "A is B" to infer "B is A".
2) Learning intersection (∩) and union (∪) operations on two sets through examples and then
    a) performing intersection and union operations on two new sets.
    b) performing intersection and union operations on three new sets ($A \cap B \cap C$ and $A \cup B \cup C$).
    c) performing a mixture of intersection and union operations on three new sets, including $(A \cap B) \cup C$ and $(A \cup B) \cap C$.

Because of the bidirectional design of the Transformer encoder, we expect that BERT should be capable of inferring "B is A" from "A is B". With both union and intersection operations being expressed sequentially (without reversal), we hypothesize that GPT models should perform at least as well as BERT, especially for tasks outlined in 2), where the deductive reasoning capabilities are needed.

**RESULTS**

***BERT model is not susceptible to "Reversal Curse"***

We have conducted a comparable experiment using the same dataset from the original reversal curse paper[5], adapted to text classification task. As shown in the **Figure 2A,** BERT model performed well on both directions and dataset types: NameToDescription and DescriptionToName. Specifically, the accuracy is around 99% when the fine-tuned model is tested on reverse direction. In the original paper[5], the accuracy of reverse direction for GPT-3-175B, is near 0. The same results hold for the newly introduced Llama 3. Given that GPT models, such as GPT-3 and Llama 3, are pretrained on extensive datasets, reversing the direction of expression typically results in outputs that are significantly shaped by this pre-existing knowledge. Consequently, the impact of fine-tuning becomes less apparent in these cases. While it is not completely fair to use accuracy as the single criterion for comparison between the decoder language model (GPT) and the encoder language model (BERT), we nonetheless include an accuracy comparison in **Figure 2B** for readers' convenience.

In summary, while it is not entirely equitable to directly compare decoder and encoder LLMs that are evaluated using different metrics, the findings do suggest that the bidirectional encoder LLM demonstrates greater proficiency in mastering reversal deduction tasks. When choosing the most appropriate LLM for different tasks, it is crucial to pick a model whose architecture matches the specific needs of the task. For example, for tasks that require reverse logical deduction, a bidirectional encoder model tends to be effective, especially when such a task can be reformulated into a classification problem.

Note that in **Figure 2**, we used GPT-3 (175B) to enable a comparison with the original "reversal curse" paper's results[5], which also utilized the GPT-3 model for their experiments. All other tables and graphs use the most updated models. However, we may switch between open-source models (Llama 2 and 3) and closed-source models (GPT-4) depending on whether fine-tuning on a curated dataset is required.

*Large language models work well in simple logical inference involving two sets*

We have conducted fine-tuning and testing on both encoder and decoder language models to verify if they can infer the Intersection and Union of two sets (we will refer to these two tasks and related dataset as the "simple" case in subsequent discussions). We found that all the BERT, Llama 2 and Llama 3 models performed well on the intersection and union on two sets, as shown in **Table 1**.

To assess the BERT model, our training dataset comprises 4000 sentences in total, with 2000 allocated for intersection and 2000 allocated for union tasks. Each task type, intersection, and union, includes 500 sentences from four distinct synthetic name groups (Superhero, Dinosaur, Mammal, and Bird). Within these 500 sentences from each group, there are equal number of positive and negative sentences, amounting to 250 each. Our train-test ratio is 10:1, meaning that there are a total of 400 sentences for testing, including 200 for intersection and 200 for union tasks. Each task type includes 50 sentences from four synthetic groups. Within these 50 sentences from each group, there are 25 positive and 25 negative sentences. **Figure 3** shows the example of testing the intersection for the cohort of "Superhero".

To evaluate the Llama 2 and 3, we exclusively utilize the positive sentences in the above-mentioned dataset for training. During inference, autoregressive Llama models focus exclusively on identifying the correct sentence rather than predicting incorrect answers. Due to the fact that there are infinitely many possible incorrect answers and only one correct answer, we do not use negative training data in our approach. It is impractical for models to deduce the correct answer by learning all the negative cases. During training, we remove the "result" segment of each sentence, allowing Llama 2 and 3 to autoregressively generate subsequent tokens. In the testing phase, Llama 2 and 3 is tasked with next token prediction until it generates the "end of sentence" token. Following this, we extract only the name entities from the output and compare the overlap with the ground truth.

As shown in **Table 1,** accuracies of Llama 2 and 3 are slightly lower than that of BERT (99%). The observed outcome could be partially attributed to the varying tasks assigned to the BERT, Llama 2 and 3 models, resulting in different metrics being used to evaluate their outcomes: The BERT model employed binary classification, with correctness determined by the label of the prediction, whereas both Llama 2 and 3 models aimed to auto-regressively predict the intersection or union of two sets by completing a prompt sentence. Nevertheless, regardless of the differing nature of their tasks, both models exhibited robust performance in simple set operations, as anticipated. Although all the BERT and the Llama models have achieved exceptionally high accuracies, it remains uncertain whether they genuinely grasp the underlying logic or merely excel at identifying patterns in the prompt's keywords. To explore this further, we will evaluate the complex logical deduction capabilities of both models.

*Large language models work poorly on complex logical inference involving three sets*

We directly applied our fine-tuned BERT, Llama 2 and 3 from the second experiment, which were trained on sentences with respect to intersection and union of two sets, to complex deductive logical inference tasks. For clarity, we use "complex" to refer to inferences involving more than two sets, in contrast to the "simple" case involving only two sets.

Specifically, our evaluation process involves four distinct scenarios for each of the four groups: Superhero, Dinosaur, Mammal, and Bird. We assess all three models' effectiveness in scenarios of A∩B∩C, A∪B∪C, (A∩B)∪C, (A∪B)∩C, using three randomly chosen sets (A, B, C) of name entities from each group. For the Llama 2 and 3 models, we create 50 positive sentences per scenario for each group, totaling 800 sentences for evaluation. This method follows the same testing approach as in the simple logical inference example, where we progressively complete a partial prompt and then evaluate the predicted entities against the actual results. For the BERT model, we also produce an additional 50 negative sentences for each scenario and group. These negative sentences, combined with the previously created positive sentences, are then assigned a label of 0 or 1 for a text classification task.

Since BERT is an encoder-based model used for classification, while Llama 2 and 3 are decoder-based autoregressive models, their evaluation metrics differ. For the BERT model, we use "Recall (accuracy)," "Precision," "Precision (outside)," "F1," and "F1 (outside)." For Llama 2 and 3, we use "Exact match" and "80% match."

Due to the autoregressive nature of Llama 2 and 3, there are no "true negative" testing data, so F1 scores are not applicable to these models. The most comparable metrics between BERT and Llama models are Recall (accuracy) and Exact match, both defined as the number of correct positive predictions divided by the number of true positive testing data.

The "Precision (outside)" in **Table 2-5** indicates that negative samples contain names outside the union of A, B, and C, providing a more meaningful evaluation of the models' bias in different testing cases and their logical reasoning capabilities. The "80% match" in **Table 2-5** means that the Llama models correctly predict at least 80% of the elements in each testing sample.

As shown in the **Table 2-5**, the BERT model demonstrates high accuracy in predicting positive cases, but this does not automatically mean it excels at conducting deductive reasoning on complex datasets. This is mainly because the problem has been simplified into a basic binary classification task. Therefore, it is important to further explore whether BERT maintains solid reasoning behind these predictions. To achieve this, we consider two categories of negative samples as mentioned before: (i) those containing only names among A, B and C, labelled as

"Precision" in **Table 2-5**, and (ii) negative samples featuring new names not part of A, B and C, labelled as "Precision (outside)" in **Table 2-5**). The same applies to the F1 scores. It can be seen that the model demonstrates a significant bias when encountering new names outside of A, B or C. When the tested output set includes names that are not part of A, B or C, the BERT model readily identifies these as incorrect. However, on the other hand, when the incorrect output is a proper subset of the target set, or the incorrect output is the superset of the target set with redundant elements from A, B, or C, the BERT model struggles to accurately ascertain its validity. In summary, the BERT model bases its decisions not on the semantic or logical meaning of the content, but rather on the recurrence of similar terms within the sentence; It tends to categorize sentences with repetitive words as positive and those with new words as negative, without considering the actual logical connections between different datasets.

For the Llama 2 and 3, we evaluated its performance as a language model by creating test prompts that are half-completed sentences, mirroring the style of the training prompts. For instance, we used prompts like "I am the friend of Batwoman, Hulk, Spider-Man, Thor. You are the friend of Thor. Our merged group of friends include", and then allowed the model to generate responses based on these incomplete prompts. Note that we treat variations in language expression as equivalent. For instance, both "Doctor Strange" and "Dr. Strange" are considered as correct answers. We discovered that the Llama models' performance varies across four distinct tasks. In the task involving the complete union of three sets, Llama 2 achieved an accuracy rate of 56% for making entirely correct predictions, and notably, it reached a 92% success rate in predicting at least 80% of the target set accurately. In the remaining three scenarios, the Llama 2 demonstrated weak performance, with the worst case showing nearly zero accuracy in predicting outcomes for the "intersect and then union" task. It appears that the Llama 2 tends to recall previously seen elements and then presents (almost) all of them in its response. This approach is only accurate when the task involves predicting the union of all three sets. The Llama 3 exhibits deductive logical abilities comparable to those of Llama 2, but with slightly enhanced performance and linguistic skills.

To thoroughly evaluate the performance of large language models (LLMs) on logical induction tasks, we also conducted experiments on the same datasets using the GPT-4 (May 2024 version). Given that GPT-4 is a proprietary model whose specific internal mechanisms remain undisclosed, our analysis is limited to assessing its performance for the purposes of heuristic evaluation. It was observed that the GPT-4 is generally capable of accurately executing most deductive logical operations involving up to three sets, despite some errors in the scenarios such as the triple intersection ($A \cap B \cap C$). Nevertheless, when dealing with more than three sets, such as seven or eight sets, the error rate tends to increase. There is potential for future improvements in deductive logic capabilities.

For the cases where the model can successfully predict the result, the model approaches the task by initially analyzing the dataset, subsequently breaking down the task into a series of steps. This methodology helps the model to tackle challenging logical induction problems by addressing simpler questions in sequence. Since we don't know how GPT-4 was implemented, we will only give explanation for its performance based on the test results. It could be that GPT-4 has undergone large amounts of training including complex logical inference tasks similar to our test data. This training likely included exposure to complex prompts and contexts, equipping the model to decompose difficult questions into simpler steps. An example of this process is illustrated in **Note S1**. Interested readers are also encouraged to conduct tests themselves.

## DISCUSSION

In our study, we examined a bidirectional LLM, BERT, and found that it is immune to the reversal curse that was reported in unidirectional LLMs such as GPT previously. We further embarked on evaluating more complex but essential sequential deductive reasoning capabilities on both encoder and decoder types of LLMs. Below we first describe an intuitive explanation of reversal curse under simple linear models, then discuss the limitation of unidirectional language models and how bidirectional models can address such limitations. Finally, we discuss how different LLMs struggle with deductive logical inference pertaining to intersection and union.

### *Reversal Curse in linear models*

It can be challenging to intuitively grasp why this the "Reversal Curse", where after learning that "A is B", one subsequently fails to deduce "B is A", occurs in intricate machine learning models. However, when using trained model weights blindly, we discuss that the "Reversal Curse" can occur even in the most basic form of supervised learning, which is simple linear regression.

To be precise, let $\Omega$ be our ground probability space (potentially very large) and let $X$ and $Y$ be two (unknown) probability distributions on $\Omega$. Let $\Omega_X \times \Omega_Y = \{x_1, \ldots, x_n\} \times \{y_1, \ldots, y_n\} \subset \Omega \times \Omega$ be set of i.i.d. training samples generated from the joint law $(X, Y)$. Assume that marginal laws of $X$ and $Y$ are very different in the sense that $\Omega_X \cap \Omega_Y = \emptyset$ almost surely. For instance, let $X$ be the distribution of names of politicians and $Y$ be the distribution of descriptions consisting of chronological order of positions hold in the country of service. After sampling, we may get $x_1$ is "Jimmy Carter", $y_1$ is "the 39[th] president of the United States", $x_2$ is "Ronald Reagan", $y_2$ is "the 40[th] president of the United States", $x_3$ is "Mikhail Gorbachev" and $y_3$ is the "the 8[th] president of the Soviet Union", etc.

Suppose we fix some embedding such that these word strings are bijectively represented as numerical values, i.e., $\Omega = \mathbb{R}$. The simplest supervised learning model we can use is obviously linear regression, which means that we assume the optimal regression function $m(x) = \mathbb{E}[Y|X = x]$ takes the form of $m(x) = \beta_0 + \beta_1 x$.

It is well-known (see e.g. the literature[34-36]) that under the assumption that the joint law of $(X, Y)$ is known and suppose

$$(\beta_0^*, \beta_1^*) := \operatorname*{argmin}_{\beta_0, \beta_1} \mathbb{E}_{(X,Y)}\left[(Y - m(X))^2\right]$$
$$= \operatorname*{argmin}_{\beta_0, \beta_1} \mathbb{E}_{(X,Y)}[(Y - \beta_0 - \beta_1 X)^2],$$

then we can prove that

$$\beta_1^* = \frac{\operatorname{Cov}(X,Y)}{\operatorname{Var}(X)} \quad \text{and} \quad \beta_0^* = \mathbb{E}[Y] - \beta_1^* \mathbb{E}[X]. \qquad \text{(Equation 1)}$$

Consequently, the *true* regression function with $X$ and $Y$ reversed takes the form of

$$X = \frac{\operatorname{Cov}(X,Y)}{\operatorname{Var}(Y)} \cdot Y + \left(\mathbb{E}[X] - \frac{\operatorname{Cov}(X,Y)}{\operatorname{Var}(Y)} \cdot \mathbb{E}[Y]\right)$$
$$= \frac{\operatorname{Cov}(X,Y)}{\operatorname{Var}(Y)} \cdot (Y - \mathbb{E}[Y]) + \mathbb{E}[X]$$

Now, suppose we simply use $\beta_0^*, \beta_1^*$ to predict the reversed logical sentences, i.e., the predicted value $X_1$ is calculated via

$$X_1 = \beta_1^* Y + \beta_0^*$$
$$= \frac{\operatorname{Cov}(X,Y)}{\operatorname{Var}(X)} \cdot Y + \left(\mathbb{E}[X] - \frac{\operatorname{Cov}(X,Y)}{\operatorname{Var}(X)} \cdot \mathbb{E}[Y]\right)$$
$$= \frac{\operatorname{Cov}(X,Y)}{\operatorname{Var}(X)} \cdot (Y - \mathbb{E}[Y]) + \mathbb{E}[X]$$

By comparing above two equations, the difference $|X_1 - X|$ could potentially be infinitely large by adjusting the variances of $X$ and $Y$. Therefore, the effect of "Reverse Curse" under the simplest linear model could be very significant. Note that the above calculation works for *any* two unknown (reasonable) distributions, with only finite second moments are assumed.

In the case of more intricate models such as deep neural networks and LLMs, it is not possible to express the estimated parameters in closed-form equations based on the statistics of unknown underlying distributions A and B. As a result, theoretical explanations for the reversal curse are limited to linear regression and cannot extend beyond it.

### *BERT can deduce "B is A" from "A is B"*

Contrasting with the GPT models, BERT is a bidirectional transformer-based encoder model. While their structures are quite similar, a key difference is that BERT does not employ a masking mechanism in its multi-head attention process. During training, all input tokens interact with each other to compute the final loss function. A classification layer is applied on top of the classification token, and binary cross entropy is used as the loss function. This bidirectional structure ensures that both ⟨name⟩ and ⟨description⟩ are updated jointly during training, irrespective of their sequence in the prompt. Once trained, the model has "learned" to use information from both directions to make inferences. This means that even if the order of A and B is reversed, their joint presence will still aid the BERT model in accurately predicting the final outcomes.

The fundamental distinction between decoder and encoder language models resides in their respective methods of managing context and sequence during the training process. GPT's unidirectional focus can limit its ability to infer reverse logical relationships, which is a result of its training mechanism that relies on the sequence of the data presented. On the other hand, BERT's bidirectional nature allows it to effectively grasp the context in both directions, making it more proficient in understanding and predicting relationships in reverse order. This fundamental difference in structure and training methodology makes BERT more versatile for tasks that require an understanding of bidirectional or reverse relationships, while GPT may excel in tasks that benefit from its sequential prediction capabilities.

In practical cases, the structural differences between GPT and BERT have significant implications for their real-world applications. For tasks involving sequence generation, storytelling, or content creation, where the flow and continuity from one token to the next are paramount, GPT's sequential predictive ability makes it a preferred choice. Its design to predict the next token based on the preceding context aligns well with these requirements. In contrast, BERT's bidirectional understanding is particularly beneficial in tasks requiring nuanced comprehension of context, such as sentiment analysis, text classification, and question-answering, where understanding the interplay between all parts of the input is crucial. This ability to process and learn from input in both directions simultaneously allows BERT to capture complex relationships and subtleties in language that might elude a unidirectional model like GPT. Therefore, the choice between GPT and BERT should be guided by the specific requirements and nature of the task at hand, leveraging their respective strengths in sequence prediction and bidirectional context comprehension.

***LLMs struggle with deductive logical inference pertaining to intersection and union***

Our study showed that both encoder (BERT) and decoder (Llama 2 and 3) language models excel in tasks involving basic intersection and union of two sets. However, they struggle with more complex logical operations that involve multiple sets.

According to **Table 1**, BERT demonstrated slightly superior performance compared to the Llama models, which could be attributed to several factors:

1) The methods used to calculate accuracy scores for these models varied, potentially impacting their comparative analysis. For the BERT model, we employed a straightforward binary classification accuracy metric. In contrast, for the Llama models, we assessed accuracy based on whether the fine-tuned model could accurately predict every name in the target set (counted as positive). If it failed to do so, the outcome was deemed negative.
2) The complexity of predictions for the target set differs between the two models. The BERT model generates two outputs indicating the likelihood of binary results. In contrast, the Llama models, specifically Llama 2, evaluate each token of a name in the target set against 32,000 possible candidates. As a result, Llama models face a larger and more challenging network for fine-tuning.
3) Considering the number of parameters that need to be trained for both models, Llama models3 likely require a more extensive dataset for fine-tuning.

To investigate further, we aimed to determine whether a language model, fine-tuned on simple intersection and union tasks involving two sets, could handle more complex set operations, such as union and intersection among three sets.

The findings, presented in **Table 2-5**, revealed that both language models exhibited constrained learning capabilities and biases in their inferences. The BERT model showed sensitivity to external words, tending to assign a negative label whenever an unfamiliar word appeared. However, when negative samples consisted solely of previously encountered words, the model struggled to accurately discern the correct label. The Llama models exhibited similar challenges. It appeared that there was no consistent pattern in how the Llama models interpreted commands. For tasks involving the union of three sets, they performed well, achieving up to 90% accuracy in predicting at least 80% of the mentioned entities. However, this high accuracy does not necessarily indicate a correct understanding of the prompt by the Llama models. This point is again disproved by its performance in the other three tasks, where the model often made numerous false positive predictions. Like the BERT model, Llama models tended to combine all words from the provided sets, overlooking the actual meaning of the command. In conclusion, we hypothesize that while both the fine-tuned BERT and Llama models can learn to perform intersection and union operations separately during training (with parameter updates through gradient descent), their inability to handle more complex operations like $A \cap (B \cup C)$ without prior exposure hints at a limitation. This limitation stems from the models' incapacity to autonomously understand the sequence and combination of operations. This finding suggests that LLMs may be more adept as writing assistants, replicating learned patterns, rather than as logical machines capable of independently deducing complex relationships or sequences. This insight into their capabilities underscores the importance of tailored training and the potential need for more advanced models or techniques to handle intricate logical tasks.

Given these findings, it is crucial to exercise a high level of caution at this stage in the development of autoregressive GPT models, particularly for applications requiring precise and sophisticated logical operations. For example, the deployment of open-source GPT models in constructing extensive medical knowledge graphs is a relevant case. Although LLMs have demonstrated their ability to assist in extracting phenotypic entities[37-40] and learning their ontological relationships[40-43] from long clinical notes so that they can benefit other downstream tasks like rare genetic diseases prediction[44], tasks involving logical reasoning may require manipulating sets of Human Phenotype Ontology (HPO) terms through operations like intersections and unions. To illustrate, if HPO term A is typically associated with 10 other HPO terms, and HPO term B is similarly associated with a different set of 10 terms, the objective is to identify the HPO terms that are commonly associated with both term A and term B. However, our research shows that LLMs are not competent in this task, let alone finding commonalities that take ontological relationships into account.

Furthermore, another challenging application involves using GPT models for executing rigorous mathematical proofs, which require a blend of complex and inventive logical operations. If these models only grasp the surface-level aspects of language and fail to fully comprehend the underlying logical structures, then the accuracy of the proofs they generate cannot be assured. Addressing this limitation would likely necessitate not only extensive domain-specific training data but also the development of more sophisticated algorithms designed to enhance the models' understanding of complex mathematical logic. These enhancements are critical for ensuring the reliability and utility of GPT models in high-stakes environments where precise logic and reasoning are paramount.

**Experimental Procedures**

***Resource availability***

<u>Lead contact</u>

Further information and requests for data should be directed to and will be fulfilled by the lead contact, Dr. Kai Wang (wangk@chop.edu).

<u>Material availability</u>

This study did not generate new unique materials.

<u>Data and code availability</u>

All the code for generating synthetic training and testing data and fine-tuning and evaluating on BERT, Llama 2 and 3 models, are made publicly accessible at https://github.com/WGLab/Reversal_Curse and have been archived at Zenodo[45].

For the evaluation of the "reversal curse" on the Llama 3 model (see **Figure 2**), we utilized the dataset available on the GitHub page (https://github.com/lukasberglund/reversal_curse/tree/main/data/reverse_experiments/june_version_7921032488) of original reversal curse study[5].

***Reversal Deduction***

<u>Data Preparation</u>

To compare with the Llama models in the published literature[5], we employed the exact same training and testing dataset from its online repository. In the original dataset, we have 30 fake pairs of ⟨name⟩ and ⟨description⟩ for training. For each pair, we include 30 different prompts in the order "⟨description⟩ to ⟨name⟩", which will help with model's generalization[46]. Similar approaches are used in the model augmentations literature as well[47-50]. For testing, the original dataset included 10 new prompts for each pair and each order, which accounts for 600 testing data points. One example of training sample is "prompt": "Known for being the renowned composer of the world's first underwater symphony, \"Abyssal Melodies.\",", "completion": " Uriah Hawthorne now enjoys a quite life." And one example of corresponding test sample (with same direction of logic) is: "prompt": "Immersed in the world of composing the world's first underwater symphony, \"Abyssal Melodies.\",", "completion": " Uriah Hawthorne". One example of corresponding test sample (with reverse direction of logic) looks like: "prompt": "The trailblazer known as Uriah Hawthorne was once", "completion": " the renowned composer of the world's first underwater symphony, \"Abyssal Melodies.\"."

Please note that the "quite" word is from the original data set literally and may be a typo for "quiet".

The original dataset was used for training a decoder model (Llama) in an autoregressive way, and the dataset is unsupervised. The results of newly introduced Llama 3 were obtained in the exact same way. For the BERT-based model, we need to make some adaptions. Firstly, we created 30 new fictitious names using ChatGPT. We then randomly substituted these 30 names into half of the original 900 positive prompts, which were selected at random. For example, the same example prompt will become "Known for being the renowned composer of the world's first underwater symphony, "Abyssal Melodies", Landon Everwood now enjoys a quite life." and "The renowned composer of the world's first underwater symphony, "Abyssal Melodies" is called Sierra Thornfield." etc. In the end, we have 900 prompts with both positive and negative labels, all of which consist of fake ⟨name⟩ and ⟨description⟩ pairs. In a similar manner, we replace the initial 30 names in the test data with our 30 newly generated names, resulting in an additional 600 negative test data points.

For a fair comparison, we also train the model on the other direction "⟨name⟩ to ⟨description⟩" but with completely different ⟨name⟩ and ⟨description⟩ to improve the meta learning of the model. The similar technique was applied in the original paper as well. In addition, we test on the other direction in the exact same pattern as "⟨description⟩ to ⟨name⟩". As a result, we ended up with a total of 1800 data points for training and 2400 for testing. We will train our BERT model as a text classification model and compute the accuracy of the model according to whether it can distinguish a positive prompt from a negative one (positive data will be assigned a label of 1, while negative data will be labeled 0).

*Fine-tuning BERT model*

We fine-tuned the vanilla pretrained BERT model (110M parameters), bert-base-cased, implemented with HuggingFace[51]. We perform hyperparameter sweeps on the model with 5, 8,

10 epochs, using batch sizes of 8, 16, 32, 64 and learning rates of 1e-06, 2e-06, 1e-05 and fix the best parameters for validation set. Finally, we conducted fine-tuning and testing of the BERT model specifically for a text classification task.

*Fine-tuning Llama 3 model*

We fine-tuned the Llama 3 model (8B parameters) with HuggingFace implementation[3] on the original reversal curse dataset (available at GitHub page [https://github.com/lukasberglund/reversal_curse/tree/main/data/reverse_experiments/june_version_7921032488](https://github.com/lukasberglund/reversal_curse/tree/main/data/reverse_experiments/june_version_7921032488)).

**Intersection and Union**

*Data Preparation*

We employed "full randomization of synthetic names" to produce training and testing datasets for executing intersection and union operations on all the BERT, Llama 2 and 3 models. Specifically, we first generate four sets of "synthetic names", each of size 30:

1) Superhero names. Sample names include "Superman", "Batman", "Iron man", etc.
2) Dinosaur names. Sample names include "Tyrannosaurus Rex", "Velociraptor", "Triceratops", etc.
3) Mammal names. Sample names include "Lion", "Tiger", "Bear", etc.
4) Bird names. Sample names include "Sparrow", "Eagle", "Owl", etc.

For each group of names X with |X| = 30, we begin by randomly selecting two integers, denoted as $s_1$ and $s_2$, between 1 and 10. Subsequently, we randomly sample two sets of names, each with size of $s_1$ and $s_2$, respectively. Denote these two sets by $S_1$ and $S_2$. Next, we use $S_1$ and $S_2$ to construct both positive and negative samples. To make our discussions clearer, there are four cases:

1. Positive samples on intersection.

Suppose

$S_1$ = [Thor, Captain America, Doctor Strange],

$S_2$ = [Black Panther],

then the data sample looks like "I am the friend of Thor, Captain America, Doctor Strange. You are the friend of Black Panther. We do not have common friends."

Suppose

$S_1$ = [Captain Marvel, Aquaman],

$S_2$ = [Captain Marvel, Daredevil, Green Arrow, Batman],

then the data sample looks like "I am the friend of Captain Marvel, Aquaman. You are the friend of Captain Marvel, Daredevil, Green Arrow, Batman. Our common friends include Captain Marvel."

2. Negative samples on intersection.

For constructing negative samples, we would randomly sample $\left[\frac{|S_1 \cap S_2|}{2}\right]$ elements (true samples) in $S_1 \cap S_2$ and then randomly sample $\left[\frac{|X \setminus (S_1 \cap S_2)|}{4}\right]$ elements in $X \setminus (S_1 \cap S_2)$ (false samples). For instance, suppose

$S_1$ = [Luke Cage, Black Widow, Iron Man],

$S_2$ = [Spider-Man, Iron Man, Hawkman, Doctor Strange, Batwoman, Luke Cage, Batman],

then the data sample looks like "I am the friend of Luke Cage, Black Widow, Iron Man. You are the friend of Spider-Man, Iron Man, Hawkman, Doctor Strange, Batwoman, Luke Cage, Batman. Our common friends include Green Lantern, Black Widow, Doctor Strange, Thor, Supergirl, Captain Marvel, Iron Man, Superman."

3. Positive samples on union.

Suppose

$S_1$ = [Thor, Captain America, Doctor Strange] and

$S_2$ = [Black Panther],

then the data sample looks like "I am the friend of Thor, Captain America, Doctor Strange. You are the friend of Black Panther. Our merged group of friends include Thor, Captain America, Doctor Strange, Black Panther."

4. Negative samples on union.

For constructing negative samples, we would randomly sample $\left[\frac{|S_1 \cup S_2|}{2}\right]$ elements (true samples) in $S_1 \cup S_2$ and then randomly sample $\left[\frac{|X \setminus (S_1 \cup S_2)|}{4}\right]$ elements in $X \setminus (S_1 \cap S_2)$ (false samples). For instance, suppose

$S_1$ = [Batwoman, Hulk, Spider-Man, Thor],

$S_2$ = [Thor],

then the data sample looks like "I am the friend of Batwoman, Hulk, Spider-Man, Thor. You are the friend of Thor. Our merged group of friends include Hawkman, Green Lantern, Thor, Black Panther, Batman, Hulk, Wolverine, Iron Man."

5. Complex testing dataset involving three new sets.

To evaluate the model's ability to make inferences, we constructed a complex dataset involving various set operations: $A \cap B \cap C$, $A \cup B \cup C$, $(A \cap B) \cup C$, $(A \cup B) \cap C$. The prompts designed for these four scenarios closely resemble the examples previously mentioned in 1-4. Specifically, we have outlined one representative (positive) prompt for each case, as detailed below:

- $A \cap B \cap C$, "Steven is the friend of Hawkman, Batwoman, Batman, Captain America, Nightwing, Superman, Green Lantern, Hulk. Frank is the friend of Batwoman, Hawkman, Storm, The Flash, The Wasp, Aquaman, Wonder Woman. Tom is the friend of Hawkman, Shazam, Hulk, Batwoman, Iron Man. The common friends between Steven, Frank and Tom include Hawkman, Batwoman."
- $A \cup B \cup C$, "Steven is the friend of Deadpool, Aquaman, Thor. Frank is the friend of Supergirl, Hulk, The Flash, The Wasp, Iron Man. Tom is the friend of Wolverine, Ant-Man, The Flash, Aquaman, Supergirl. Their merged group of friends include Ant-Man, Supergirl, Wolverine, Hulk, The Flash, Thor, The Wasp, Iron Man, Deadpool, Aquaman."
- $(A \cap B) \cup C$, "Steven is the friend of Cyclops, Green Lantern. Frank is the friend of Luke Cage, Cyclops, The Flash. Tom is the friend of Daredevil, Green Lantern, Storm, Green Arrow. The merged group of friends that include the Steven and Frank's common friends and friends of Tom include Cyclops, Green Arrow, Green Lantern, Storm, Daredevil."
- $(A \cup B) \cap C$, "Steven is the friend of Luke Cage, Ant-Man, Scarlet Witch. Frank is the friend of Green Lantern, Hulk. Tom is the friend of Shazam, Black Widow, Hulk, Black Panther, Storm, Luke Cage, The Wasp, Spider-Man. The common friends between Steven and Frank's merged group of friends and friends of Tom include Hulk, Luke Cage."

Additionally, we created corresponding negative prompts for each scenario in a similar manner to before. It's important to note that all these negative examples were exclusively used for the BERT model. Since our goal is to specifically assess the logical deduction capabilities of GPT models, we standardized the language template to "is friend of" during both training and testing. This template could be substituted with other phrases as long as consistency is maintained when comparing across different language models. We experimented with various terms to describe set properties (such as "colleagues"), but these changes do not impact the logical deduction tests (such as union or inclusion) themselves.

### *Fine-tuning on BERT and Llama models*

We fine-tuned both the BERT model (bidirectional encoder) and the open-sourced Llama 2 and 3 (unidirectional decoder) to compare their performance on the "Intersection and Union" task. For training, we employed only simple datasets involving relationships between two sets, while for testing, we used both simple and complex datasets (involving three sets).

For the BERT model, similar to the reversal deduction process, we fine-tuned the pretrained "bert-base-cased" model for a text classification task. This was done using the positive and negative simple datasets. Furthermore, we reserved 10% of this data (pertaining to either intersection or union scenarios) as previously mentioned, only for validation purposes. The accuracy of the testing was determined based on the results of binary classification with the fine-tuned BERT model.

For the Llama 2 and 3, we employed the LoRA[52] training approach to enhance training efficiency while maintaining the model's performance. Since we will fine-tune Llama 2 and 3 as generative language models (next token prediction), we do not need negative samples in this unsupervised training approach. We fine-tuned the model exclusively on positive samples from simple datasets (covering both intersection and union), with 10% of this data set aside for validation purposes. In the test dataset, we deleted the target output from the prompt and allowed the fine-tuned model to autonomously complete it (e.g. "… Our merged group of friends include Ant-Man, Supergirl, Wolverine, Hulk, The Flash, Thor, The Wasp, Iron Man, Deadpool, Aquaman."). We then extracted the predicted sets and compared their coverage with the target dataset. To assess correctness, we employed two thresholds: a perfect match and an 80% match.

To comprehensively assess the logical inference capabilities of LLMs, we evaluated the previously fine-tuned models, BERT, Llama 2 and 3, which had been trained with simple datasets, against complex datasets as well. The testing process is exactly same as above.

## ACKNOWLEDGEMENTS

We thank members of the Wang lab for comments and feedback on the analytical procedures. The study is supported by NIH grant HG013031 and the CHOP Research Institute.

## AUTHOR CONTRIBUTIONS

JY and DW designed the experimental procedures and performed data analysis. KW supervised the study. All authors contributed to the writing of the manuscript.

## DECLARATION OF INTERESTS

The authors declare no competing interests.

## REFERENCES

1. Devlin, J., Chang, M.-W., Lee, K., and Toutanova, K. (2018). Bert: Pre-training of deep bidirectional transformers for language understanding. Preprint at arXiv. https://doi.org/10.48550/arXiv.1810.04805.
2. Achiam, J., Adler, S., Agarwal, S., Ahmad, L., Akkaya, I., Aleman, F.L., Almeida, D., Altenschmidt, J., Altman, S., and Anadkat, S. (2023). GPT-4 technical report. Preprint at arXiv. https://doi.org/10.48550/arXiv.2303.08774.
3. AI@Meta (2024). Llama 3 Model Card. https://github.com/meta-llama/llama3/blob/main/MODEL_CARD.md.

4. Floridi, L., and Chiriatti, M. (2020). GPT-3: Its nature, scope, limits, and consequences. Minds and Machines *30*, 681-694.

5. Berglund, L., Tong, M., Kaufmann, M., Balesni, M., Stickland, A.C., Korbak, T., and Evans, O. (2023). The Reversal Curse: LLMs trained on" A is B" fail to learn" B is A". Preprint at arXiv. https://doi.org/10.48550/arXiv.2309.12288.

6. Brown, T., Mann, B., Ryder, N., Subbiah, M., Kaplan, J.D., Dhariwal, P., Neelakantan, A., Shyam, P., Sastry, G., and Askell, A. (2020). Language models are few-shot learners. Advances in neural information processing systems *33*, 1877-1901.

7. Touvron, H., Lavril, T., Izacard, G., Martinet, X., Lachaux, M.-A., Lacroix, T., Rozière, B., Goyal, N., Hambro, E., and Azhar, F. (2023). Llama: Open and efficient foundation language models. Preprint at arXiv. https://doi.org/10.48550/arXiv.2302.13971.

8. Grosse, R., Bae, J., Anil, C., Elhage, N., Tamkin, A., Tajdini, A., Steiner, B., Li, D., Durmus, E., and Perez, E. (2023). Studying large language model generalization with influence functions. Preprint at arXiv. https://doi.org/10.48550/arXiv.2308.03296.

9. Meng, K., Bau, D., Andonian, A., and Belinkov, Y. (2022). Locating and editing factual associations in GPT. Advances in Neural Information Processing Systems *35*, 17359-17372.

10. St Clair-Thompson, H.L., and Allen, R.J. (2013). Are forward and backward recall the same? A dual-task study of digit recall. Memory & cognition *41*, 519-532.

11. Thomas, J.G., Milner, H.R., and Haberlandt, K.F. (2003). Forward and backward recall: Different response time patterns, same retrieval order. Psychological Science *14*, 169-174.

12. Bireta, T.J., Fry, S.E., Jalbert, A., Neath, I., Surprenant, A.M., Tehan, G., and Tolan, G.A. (2010). Backward recall and benchmark effects of working memory. Memory & cognition *38*, 279-291.

13. Li, S.-C., and Lewandowsky, S. (1995). Forward and backward recall: Different retrieval processes. Journal of Experimental Psychology: Learning, Memory, and Cognition *21*, 837.

14. Guitard, D., Saint-Aubin, J., Poirier, M., Miller, L.M., and Tolan, A. (2020). Forward and backward recall: Different visuospatial processes when you know what's coming. Memory & Cognition *48*, 111-126.

15. Geva, M., Schuster, R., Berant, J., and Levy, O. (2020). Transformer feed-forward layers are key-value memories. Preprint at arXiv. https://doi.org/10.48550/arXiv.2012.14913.

16. Geva, M., Caciularu, A., Wang, K.R., and Goldberg, Y. (2022). Transformer feed-forward layers build predictions by promoting concepts in the vocabulary space. Preprint at arXiv. https://doi.org/10.48550/arXiv.2203.14680.

17. Geva, M., Bastings, J., Filippova, K., and Globerson, A. (2023). Dissecting recall of factual associations in auto-regressive language models. Preprint at arXiv. https://doi.org/10.48550/arXiv.2304.14767.

18. Vaswani, A., Shazeer, N., Parmar, N., Uszkoreit, J., Jones, L., Gomez, A.N., Kaiser, Ł., and Polosukhin, I. (2017). Attention is all you need. Advances in neural information processing systems *30*.

19. Mizrahi, M., Kaplan, G., Malkin, D., Dror, R., Shahaf, D., and Stanovsky, G. (2023). State of what art? a call for multi-prompt llm evaluation. Preprint at arXiv. https://doi.org/10.48550/arXiv.2401.00595.

20. Wang, B., Yue, X., and Sun, H. (2023). Can ChatGPT defend its belief in truth? evaluating LLM reasoning via debate. Findings of the Association for Computational Linguistics: EMNLP 2023, 11865-11881.

21. Eigner, E., and Händler, T. (2024). Determinants of LLM-assisted Decision-Making. Preprint at arXiv. https://doi.org/10.48550/arXiv.2402.17385.

22. Zhang, R., Jiang, D., Zhang, Y., Lin, H., Guo, Z., Qiu, P., Zhou, A., Lu, P., Chang, K.-W., and Gao, P. (2024). MathVerse: Does Your Multi-modal LLM Truly See the Diagrams in Visual Math Problems? Preprint at arXiv. https://doi.org/10.48550/arXiv.2403.14624.

23. Ozturkler, B., Malkin, N., Wang, Z., and Jojic, N. (2023). Thinksum: Probabilistic reasoning over sets using large language models. Proceedings of the 61st Annual Meeting of the Association for Computational Linguistics (Volume 1: Long Papers), 1216-1239.

24. Chang, E.Y. (2024). SocraSynth: Multi-LLM Reasoning with Conditional Statistics. Preprint at arXiv. https://doi.org/10.48550/arXiv.2402.06634.

25. Quan, X., Valentino, M., Dennis, L.A., and Freitas, A. (2024). Verification and Refinement of Natural Language Explanations through LLM-Symbolic Theorem Proving. Preprint at arXiv. https://doi.org/10.48550/arXiv.2405.01379.

26. Wu, D., Yang, J., Ahsan, M.U., and Wang, K. (2023). Classification of integers based on residue classes via modern deep learning algorithms. Patterns *4*.

27. Kim, J., Kovach, M., Lee, K.-M., Shin, E., and Tzavellas, H. (2024). Learning to be Homo Economicus: Can an LLM Learn Preferences from Choice. Preprint at arXiv. https://doi.org/10.48550/arXiv.2401.07345.

28. Cui, J., Li, Z., Yan, Y., Chen, B., and Yuan, L. (2023). Chatlaw: Open-source legal large language model with integrated external knowledge bases. Preprint at arXiv. https://doi.org/10.48550/arXiv.2306.16092.

29. Thorne, S. (2023). Experimenting with ChatGPT for Spreadsheet Formula Generation: Evidence of Risk in AI Generated Spreadsheets. Preprint at arXiv. https://doi.org/10.48550/arXiv.2309.00095.

30. Cheng, H., Sheng, B., Lee, A., Chaudhary, V., Atanasov, A.G., Liu, N., Qiu, Y., Wong, T.Y., Tham, Y.-C., and Zheng, Y.-F. (2024). Have AI-Generated Texts from LLM Infiltrated the Realm of Scientific Writing? A Large-Scale Analysis of Preprint Platforms. Preprint at bioRxiv. https://doi.org/10.1101/2024.03.25.586710.

31. Touvron, H., Martin, L., Stone, K., Albert, P., Almahairi, A., Babaei, Y., Bashlykov, N., Batra, S., Bhargava, P., and Bhosale, S. (2023). Llama 2: Open foundation and fine-tuned chat models. Preprint at arXiv. https://doi.org/10.48550/arXiv.2307.09288.

32. Fei, H., Ren, Y., Zhang, Y., Ji, D., and Liang, X. (2021). Enriching contextualized language model from knowledge graph for biomedical information extraction. Briefings in bioinformatics *22*, bbaa110.

33. He, B., Zhou, D., Xiao, J., Liu, Q., Yuan, N.J., and Xu, T. (2019). Integrating graph contextualized knowledge into pre-trained language models. Preprint at arXiv. https://doi.org/10.48550/arXiv.1912.00147.

34. Zou, K.H., Tuncali, K., and Silverman, S.G. (2003). Correlation and simple linear regression. Radiology *227*, 617-628.

35. Maulud, D., and Abdulazeez, A.M. (2020). A review on linear regression comprehensive in machine learning. Journal of Applied Science and Technology Trends *1*, 140-147.

36. Poole, M.A., and O'Farrell, P.N. (1971). The assumptions of the linear regression model. Transactions of the Institute of British Geographers, 145-158.

37. Yang, J., Liu, C., Deng, W., Wu, D., Weng, C., Zhou, Y., and Wang, K. (2024). Enhancing phenotype recognition in clinical notes using large language models: PhenoBCBERT and PhenoGPT. Patterns *5*.

38. Luo, L., Yan, S., Lai, P.-T., Veltri, D., Oler, A., Xirasagar, S., Ghosh, R., Similuk, M., Robinson, P.N., and Lu, Z. (2021). PhenoTagger: a hybrid method for phenotype concept recognition using human phenotype ontology. Bioinformatics *37*, 1884-1890.

39. Feng, Y., Qi, L., and Tian, W. (2022). PhenoBERT: a combined deep learning method for automated recognition of human phenotype ontology. IEEE/ACM Transactions on Computational Biology and Bioinformatics *20*, 1269-1277.

40. Cao, L., Sun, J., and Cross, A. (2024). AutoRD: An Automatic and End-to-End System for Rare Disease Knowledge Graph Construction Based on Ontologies-enhanced Large Language Models. Preprint at arXiv. https://doi.org/10.48550/arXiv.2403.00953.

41. Fisher, H.M., Hoehndorf, R., Bazelato, B.S., Dadras, S.S., King, L.E., Gkoutos, G.V., Sundberg, J.P., and Schofield, P.N. (2016). DermO; an ontology for the description of dermatologic disease. Journal of Biomedical Semantics *7*, 1-9.

42. Dong, H., Chen, J., He, Y., and Horrocks, I. (2023). Ontology enrichment from texts: A biomedical dataset for concept discovery and placement. Proceedings of the 32nd ACM International Conference on Information and Knowledge Management, 5316-5320.

43. Caufield, J.H., Hegde, H., Emonet, V., Harris, N.L., Joachimiak, M.P., Matentzoglu, N., Kim, H., Moxon, S., Reese, J.T., and Haendel, M.A. (2024). Structured prompt interrogation and recursive extraction of semantics (SPIRES): A method for populating knowledge bases using zero-shot learning. Bioinformatics *40*, btae104.

44. Wu, D., Yang, J., Liu, C., Hsieh, T.-C., Marchi, E., Blair, J., Krawitz, P., Weng, C., Chung, W., and Lyon, G.J. (2024). GestaltMML: Enhancing Rare Genetic Disease Diagnosis through Multimodal Machine Learning Combining Facial Images and Clinical Texts. Preprint at arXiv. https://doi.org/10.48550/arXiv.2312.15320.

45. Wu, D., Yang, J., and Wang, K. (2024). Code, datasets, and results for the paper "Exploring the Reversal Curse and Other Deductive Logical Reasoning in BERT and GPT-Based Large Language Models". Zenodo. https://doi.org/10.5281/zenodo.12144019.

46. Berglund, L., Stickland, A.C., Balesni, M., Kaufmann, M., Tong, M., Korbak, T., Kokotajlo, D., and Evans, O. (2023). Taken out of context: On measuring situational awareness in LLMs. Preprint at arXiv. https://doi.org/10.48550/arXiv.2309.00667.

47. Sennrich, R., Haddow, B., and Birch, A. (2015). Improving neural machine translation models with monolingual data. Preprint at arXiv. https://doi.org/10.48550/arXiv.1511.06709.

48. Cai, H., Chen, H., Song, Y., Zhang, C., Zhao, X., and Yin, D. (2020). Data manipulation: Towards effective instance learning for neural dialogue generation via learning to augment and reweight. Preprint at arXiv. https://doi.org/10.48550/arXiv.2004.02594.

49. Kobayashi, S. (2018). Contextual augmentation: Data augmentation by words with paradigmatic relations. Preprint at arXiv. https://doi.org/10.48550/arXiv.1805.06201.

50. Eldan, R., and Li, Y. (2023). Tinystories: How small can language models be and still speak coherent english? Preprint at arXiv. https://doi.org/10.48550/arXiv.2305.07759.

51. Wolf, T., Debut, L., Sanh, V., Chaumond, J., Delangue, C., Moi, A., Cistac, P., Rault, T., Louf, R., and Funtowicz, M. (2019). Huggingface's transformers: State-of-the-art natural language processing. Preprint at arXiv. https://doi.org/10.48550/arXiv.1910.03771.

52. Hu, E.J., Shen, Y., Wallis, P., Allen-Zhu, Z., Li, Y., Wang, S., Wang, L., and Chen, W. (2021). Lora: Low-rank adaptation of large language models. Preprint at arXiv. https://doi.org/10.48550/arXiv.2106.09685.

**Figure titles and legends**

**Figure 1. Reversal Curse of GPT model.** GPT-4 accurately identifies Leonhard Euler's mother when her son's name is provided. However, it struggles to recognize "Leonhard Euler" when given the name of his mother. With in-context learning and logical deduction, it correctly addressed the question.

**Figure 2. Comparison of BERT (110M), GPT-3 (175B) and Llama 3 (8B) on the "Reversal Curse."** (A) when provided with DescriptionToName (D2N) data, all the GPT-3, Llama 3 and BERT models can accurately respond to questions in the same direction (D2N same). However, while BERT continues to accurately predict outcomes for reverse D2N questions (D2N reverse), both GPT-3 and Llama 3 struggles with these reversed queries. (B) Both GPT-3 and Llama 3 (8B) have poor performance on the reversed testing examples, while BERT overcomes the reversal curse in the testing examples. Note that these two models have different metric to compute the accuracy, and for GPT and Llama 3 models, "description" is much harder to predict than "name" due to its length and diffusive variation.

D2N same/reverse: Trained on D2N sentences and tested on the same/reverse direction w.r.t. training data.

N2D same/reverse: Trained on N2D sentences and tested on the same/reverse direction

**Figure 3. Logical inference as set operation deductions.** We trained BERT, Llama 2 and 3 models on simple logical inference involving only two sets and tested them on both simple and complex (involving three sets) scenarios.

TABLES

|  | BERT | Llama 2 | Llama 3 |
|---|---|---|---|
| **Intersection** | 0.99 ± 0.00 | 0.95 ± 0.01 | 0.97 ± 0.01 |
| **Union** | 0.99 ± 0.00 | 0.96 ± 0.01 | 0.97 ± 0.01 |

**Table 1. Results for the BERT, Llama 2 and 3 in the test involving intersection and union of two sets.** Average exact entity-match accuracy (± SD) across 4 different cohorts of names. The training (resp. testing) size is 500 (resp. 50) for each scenario (intersection and union) and cohort. The BERT model's average accuracy was slightly higher than that of both Llama 2 and 3, yet both models demonstrated strong performance in operations involving two sets.

|  | BERT | | | | | Llama 2 | | Llama 3 | |
|---|---|---|---|---|---|---|---|---|---|
| **Cohort** | Recall (accuracy) | Precision | Precision (outside) | F1 | F1 (outside) | Exact match | 80% match | Exact match | 80% match |
| **Superhero** | 1.00 | 0.54 | 1.00 | 0.70 | 1.00 | 0.58 | 0.92 | 0.60 | 0.95 |
| **Dinosaur** | 0.96 | 0.52 | 0.98 | 0.67 | 0.97 | 0.50 | 0.90 | 0.58 | 0.92 |
| **Mammal** | 1.00 | 0.54 | 0.94 | 0.70 | 0.97 | 0.60 | 0.92 | 0.58 | 0.92 |
| **Bird** | 0.98 | 0.53 | 0.96 | 0.69 | 0.97 | 0.52 | 0.94 | 0.56 | 0.95 |

**Table 2. Results for the BERT, Llama 2 and 3 in the test of $A \cup B \cup C$.** "Precision (outside)" means that negative samples contain names outside the union of A, B and C. "F1 (outside)" means that the F1 score between Recall and Precision(outside).

|  | BERT | | | | | Llama 2 | | Llama 3 | |
| --- | --- | --- | --- | --- | --- | --- | --- | --- | --- |
| Cohort | Recall (accuracy) | Precision | Precision (outside) | F1 | F1 (outside) | Exact match | 80% match | Exact match | 80% match |
| **Superhero** | 0.94 | 0.55 | 1.00 | 0.69 | 0.97 | 0.07 | 0.08 | 0.10 | 0.12 |
| **Dinosaur** | 0.96 | 0.54 | 0.98 | 0.69 | 0.97 | 0.10 | 0.12 | 0.10 | 0.12 |
| **Mammal** | 1.00 | 0.56 | 0.93 | 0.72 | 0.96 | 0.14 | 0.14 | 0.11 | 0.15 |
| **Bird** | 0.96 | 0.53 | 0.96 | 0.68 | 0.96 | 0.08 | 0.10 | 0.09 | 0.10 |

**Table 3. Results for the BERT, Llama 2 and 3 in the test of $A \cap B \cap C$.** "Precision (outside)" means that negative samples contain names outside the union of A, B and C. "F1(outside)" means that the F1 score between Recall and Precision(outside).

|  | BERT | | | | | Llama 2 | | Llama 3 | |
| --- | --- | --- | --- | --- | --- | --- | --- | --- | --- |
| Cohort | Recall (accuracy) | Precision | Precision (outside) | F1 | F1 (outside) | Exact match | 80% match | Exact match | 80% match |
| Superhero | 0.95 | 0.51 | 0.96 | 0.66 | 0.95 | 0.17 | 0.17 | 0.16 | 0.17 |
| Dinosaur | 0.98 | 0.50 | 0.95 | 0.66 | 0.96 | 0.18 | 0.20 | 0.16 | 0.17 |
| Mammal | 0.96 | 0.52 | 0.98 | 0.67 | 0.97 | 0.14 | 0.16 | 0.13 | 0.18 |
| Bird | 1.00 | 0.52 | 0.96 | 0.68 | 0.98 | 0.10 | 0.12 | 0.14 | 0.13 |

**Table 4**. **Results for the BERT, Llama 2 and 3 model in the test of $(A \cup B) \cap C$.** "Precision (outside)" means that negative samples contain names outside the union of A, B and C. "F1(outside)" means that the F1 score between Recall and Precision(outside).

|  | BERT | | | | | Llama 2 | | Llama 3 | |
| --- | --- | --- | --- | --- | --- | --- | --- | --- | --- |
| Cohort | Recall (accuracy) | Precision | Precision (outside) | F1 | F1 (outside) | Exact match | 80% match | Exact match | 80% match |
| Superhero | 0.85 | 0.54 | 0.98 | 0.66 | 0.91 | 0.00 | 0.03 | 0.02 | 0.03 |
| Dinosaur | 0.90 | 0.53 | 0.96 | 0.67 | 0.93 | 0.02 | 0.03 | 0.02 | 0.03 |
| Mammal | 0.87 | 0.51 | 0.98 | 0.64 | 0.92 | 0.04 | 0.06 | 0.02 | 0.04 |
| Bird | 0.86 | 0.51 | 0.93 | 0.64 | 0.89 | 0.02 | 0.02 | 0.02 | 0.03 |

**Table 5**. **Results for the BERT, Llama 2 and 3 in the test of $(A \cap B) \cup C$.** "Precision (outside)" means that negative samples contain names outside the union of A, B and C. "F1(outside)" means that the F1 score between Recall and Precision(outside).